\documentclass[pmlr]{jmlr}

\usepackage{longtable}

\usepackage{booktabs}
\usepackage[load-configurations=version-1]{siunitx} 
\usepackage{latexsym}
\usepackage{amsmath}
\usepackage{algorithm}
\usepackage[noend]{algpseudocode}
\usepackage{multirow}
\usepackage{array}
\usepackage{float}
\usepackage{hyperref}
\usepackage{adjustbox}
\usepackage[T1]{fontenc}
\usepackage[utf8]{inputenc}
\usepackage{microtype}

\makeatletter
\def\set@curr@file#1{\def\@curr@file{#1}} 
\makeatother

\newcommand{\gptthree}{GPT-3~}
\newcommand{\gptthreecf}{GPT-3-ENS~}
\newcommand{\pegasus}{PEGASUS~}
\newcommand{\twompgen}{DRSUM~}

\DeclareMathOperator*{\argmax}{arg\,max}

\theorembodyfont{\upshape}
\theoremheaderfont{\scshape}
\theorempostheader{:}
\theoremsep{\newline}

\jmlrvolume{126}
\jmlryear{2021}
\jmlrworkshop{Machine Learning for Healthcare}

\title{Medically Aware GPT-3 as a Data Generator for Medical Dialogue Summarization} 

\author{\Name{Bharath Chintagunta}
       \Email{jaic4@vt.edu}
       \AND
       \Name{Namit Katariya}
      \Email{namit@curai.com}
       \AND
       \Name{Xavier Amatriain}
       \Email{xavier@curai.com}
       \AND
       \Name{Anitha Kannan}
       \Email{anitha@curai.com}
} 

\begin{document}

\maketitle

\begin{abstract}
In medical dialogue summarization, summaries must be coherent and must capture all the medically relevant information in the dialogue. However, learning effective models for summarization require large amounts of  labeled data which is especially hard to obtain.  We present an algorithm to create synthetic training data with an explicit focus on capturing medically relevant information. We utilize GPT-3 as the backbone of our algorithm and scale 210 human labeled examples to yield results comparable to using 6400 human labeled examples ($\sim$30x) leveraging low-shot learning and an ensemble method. In detailed experiments, we show that this approach produces high quality training data that can further be combined with human labeled data to get summaries that are strongly preferable to those produced by models trained on human data alone both in terms of medical accuracy and coherency.
\end{abstract}
\interfootnotelinepenalty=10000

\section{Introduction}

With increasing usage of telehealth platforms \cite{telemedicine}, large scale ecosystems of providers and patients have become apparent. This has exacerbated the need for comprehensive visit summaries of the medical dialogues by the attending practitioner in order to facilitate accurate hand-offs to other care providers or as a means of recording the interaction. Unfortunately, having providers write summaries after each encounter is not only time consuming but also costly, limiting the scalability of telehealth platforms
\cite{physicianburnout}

In these settings, an automated summarizer that can assist the practitioners can be extremely valuable. However, an important challenge of end-to-end medical dialogue summarization is the lack of large scale annotated datasets. Annotation of medical dialogues is expensive and slow because they need to be curated by trained experts. This is further compounded by the fact that labeled data may not be publicly shared because of patient privacy concerns and HIPAA regulations. 

Recent approaches to summarization \cite{pointergen, qi2020prophetnet, zhang2019pegasus} use transfer learning where a pre-trained model (e.g. through self supervision of learning a language model) is fine tuned with a labeled dataset. However, fine-tuning still requires hundreds to thousands of labeled examples to obtain reasonable performance.  Methods such as \cite{joshi2020dr} aim to partially overcome these issues through modeling strategies that directly learn important inductive biases from smaller amounts of data. In addition, \cite{joshi2020dr} also handled data sparsity by leveraging a key insight of sequential nature of information flow in a medical dialogue: global summary of the dialogue can be composed from local dialogue turns (snippets). This enables collecting training data for snippets as opposed to the full conversation - an insight, we use in our paper as well.

\begin{figure}[H]
\centering
\includegraphics[scale=0.5]{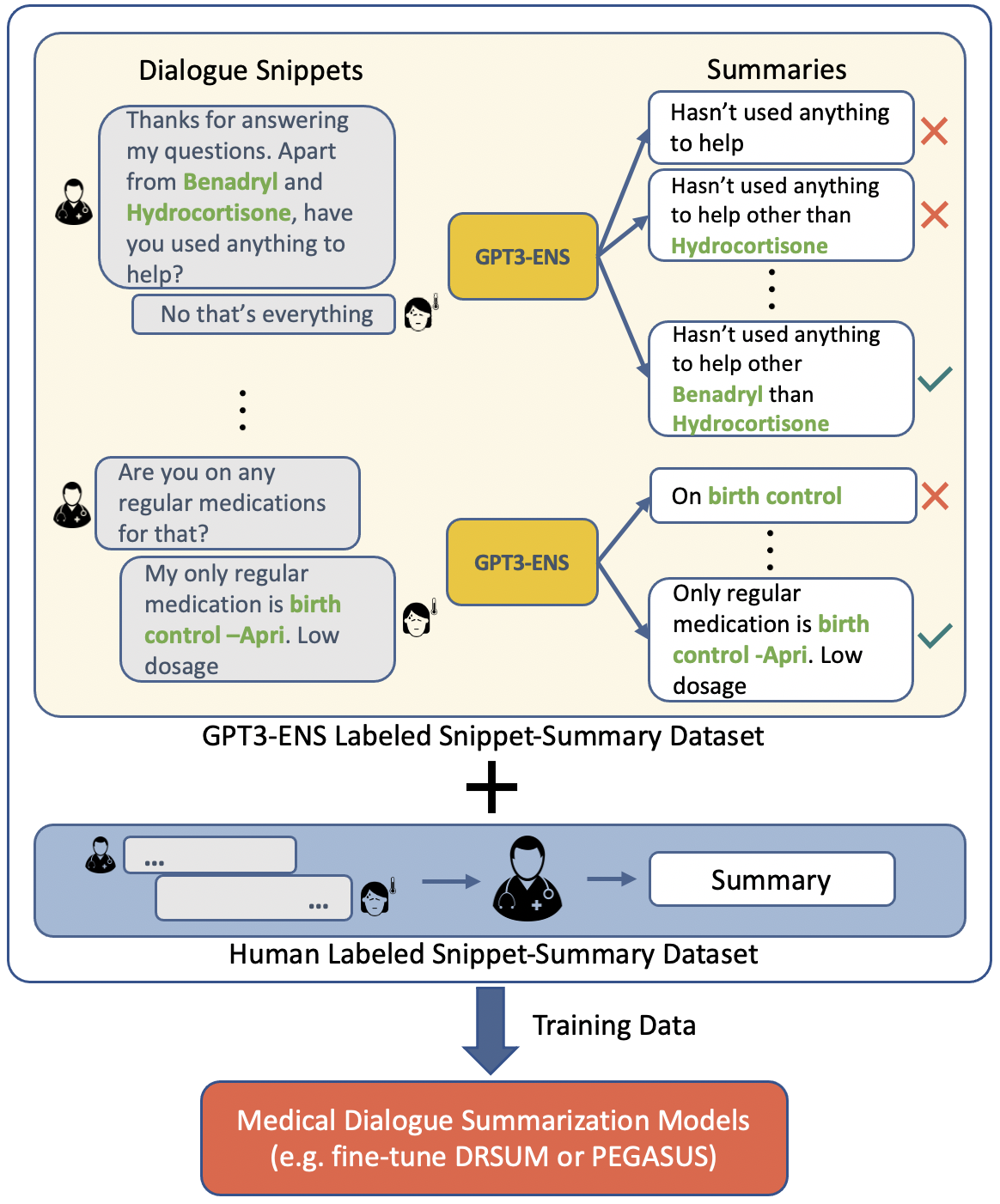}
\caption{Overview of our proposed approach: we train models on a mix of \gptthreecf synthesized and human labeled data to get performance better than models trained on either of the sources}
\label{fig:intro}
\end{figure}

Recently, OpenAI developed GPT-3, a neural language model that is capable of  natural language generation and completion of tasks like classification, question-answering, and summarization \cite{brown2020language}. The focus of that work is to enable task-agnostic and zero-shot or low-shot performance as opposed to a pre-trained model that needs to be fine-tuned separately on every downstream task. In this paper, we investigate the following question:
{\it How can a low-shot learner such as GPT-3 be leveraged to scale training data for medical dialogue summarization models?}
In answering this question within the context of GPT-3 as a black box proprietary API\footnote{\url{https://beta.openai.com/}}, we took into account multiple considerations:
\begin{itemize}
    \item \textit{Medical Correctness \cite{joshi2020dr}}:  Medical summarization warrants high precision and therefore the summarizer should be good at capturing all the medical information (medications, symptoms etc) discussed in the dialogue and (2) discern all the affirmatives and negatives on medical conditions correctly (e.g. no allergies, having a cough for 2 days). 
    \item \textit{Privacy Concerns}: At inference time, an API call to external services such GPT-3 may not always be possible due to HIPAA and privacy concerns.
    \item \textit{Practitioner in the loop}: The technique needs to be easily amenable to a feedback loop that allows for leveraging manually curated human labels. This feedback loop is extremely important because the diversity and the long tail of data distribution in medical dialogue means that there can be parts of the summary that needs to be edited by practitioners for medical correctness and completeness. Note that these edits can be used as additional data for improving the underlying model. 
\end{itemize}
Taking into account these considerations, this paper makes the following contributions (\autoref{fig:intro} for a quick overview):
\begin{itemize}
    \item  We introduce a medically-aware GPT-3 data labeler, {\gptthreecf}, that combines medical knowledge and an ensemble of GPT-3 for the purpose of medical dialogue summarization.
    \item  We introduce the idea of using \gptthreecf as a dataset generator to facilitate learning an in-house summarization model. Our experiments show that we can obtain the same performance as that of human labeled dataset with 30x smaller amount of human labeled data. With only 210 expert curated summaries and GPT-3 as a labeled data simulator, we can mimic the performance of a summarization model trained on 6400 expert curated summaries.
    \item By combining generated datasets from \gptthreecf with a human labeled dataset, we show that we can obtain better performance than models trained on either one of the data sources. 
\end{itemize}

The rest of the paper is structured as follows: \autoref{sec:related} discusses related work, \autoref{sec:gpt3cf} introduces our approach, \autoref{sec:datasets} and \autoref{sec:eval} describe our datasets and metrics respectively while \autoref{sec:expt} illustrates our experiments. We end the paper with \autoref{sec:conclusion} discussing our conclusions and future work.

\section{Generalizable Insights about Machine Learning in the Context of Healthcare}
One of the main challenges in using deep learning for healthcare is the lack of large annotated datasets. It is usually costly and time-consuming to collect a large labeled dataset because annotations need to be provided by trained healthcare professionals. As deep models usually require a large amount of data to perform accurately and robustly, this deters their widespread application in healthcare. So, it is essential to develop low-shot models in healthcare i.e. models that can do well given a small number of labeled examples. In parallel, there has been a lot of progress in development of large scale models leveraging web-scale data, such as GPT-3, that show good low-shot performance. However, these models can be noisy, particularly in the medical domain, so we need approaches that mitigate this noise but are still able to leverage these models' strengths. In this context, our approach of infusing medical knowledge in pretrained models such as GPT-3 to generate high-quality synthetic labels is an idea with wide applicability in low-resource settings like healthcare.

If pretrained models can be used to generate accurate labels, can they be directly leveraged for the task at hand? In many settings, they probably can but particularly in healthcare, this is nuanced. ML models in healthcare can learn and improve over time only if they are amenable to feedback loops i.e. they can be retrained with labels that are corrected / edited by medical practitioners. Moreover, if the model making the predictions is owned by a third party, privacy protocols (e.g. HIPAA) mandate that either they obey the same privacy protocols or that data be deidentified before being sent to such external services. Both these necessitate the need for a different approach. We introduce one such approach where we infuse medical knowledge into an external non-HIPAA compliant model (GPT-3) and leverage it as a data generator to obtain a large training set, to then train an in-house model. Since the data exposed to GPT-3 is fixed and small (in our experiments, GPT-3 only saw 210 examples), it can be ensured to be privacy protected. Our proposed approach to develop an in-house model has two advantages (1) It can be used at inference time without the practical constraint of data de-identification and (2) It lends itself well to the aforementioned practitioner-in-the-loop setting.

\section{Related work}
\label{sec:related}
{\noindent \bf Summarization} Emergence of sequence to sequence models and attention mechanisms \cite{seq2seq} has led to rapid progress on extractive \cite{extract} , abstractive \cite{cnndm2,zhang2019pegasus} and hybrid models \cite{pointergen, copynet} for summarization. Much of the recent work has shown these models to generate near-human coherent summaries while retaining reasonable factual correctness. 

\noindent {\bf Dialogue summarization}: While most neural summarization has focused on news corpora, recent work has tried to tackle unique challenges associated with summarizing dialogues. \cite{gated} proposes using dialogue history encoders based on the type of dialogue section to inform the generation. \cite{didi} propose using key points as a means of categorizing sections of dialogue.

{\noindent \bf Medical dialogue summarization} Existing work \cite{extractive,radiology,topicaware,soap,krishna2020extracting,joshi2020dr} in this space focuses on effective summarization by incorporating medical knowledge from a modeling perspective. Our work also focuses on incorporating medical knowledge from a data labeling perspective. We show how we leverage pretrained language models and low-shot learning \cite{brown2020language} to collect labeled data for medical dialogue summarization. We also show how this data can improve performance over models that are trained solely on existing human labeled data.


\newcommand{\myindent}[1]{\newline\makebox[#1cm]{}}

\section{Background: Can GPT-3 serve as a  medical summarizer?}
\label{sec:background}
Ignoring the privacy concerns and practitioner-in-the-loop considerations, we first explore whether GPT-3 \cite{brown2020language} is a good medical summarizer by itself. 

GPT-3 takes as input a {\it priming context} to perform the task on a previously unseen example. Priming context refers to the text description of a task and a few demonstrations of the task being accomplished (in our case, that would be dialogue snippet summarization). 

Table~\ref{tab:examples} column 2 provides examples of summaries generated by the GPT-3 model. We can clearly see that it misses a number of important pieces of information in the snippets -- first, missing medical concepts making the summary unusable (Rows 1-2). Second, the model may not always get the affirmations correct (Row 3). Third, the summary may repeat redundant information from the doctor's queries (Row 4).

Based on these observations, we might {\it prematurely} conclude that GPT-3 can not be used for medical summarization task. However, our key observation in exploring GPT-3 is that it is sensitive to the priming context (also reported in \cite{liu2021makes}), as the model does not learn but just adheres to the examples given. As we show in \ref{sec:gpt3cf}, we exploit this variability in GPT-3 output via ensembling and infusion of medical knowledge so that it can be used as a part of an effective low-shot learning approach to medical summarization.  

\begin{table}[ht!]
\footnotesize
\renewcommand{\arraystretch}{1.3}

\begin{tabular}{p{0.3\textwidth}m{0.3\textwidth}m{0.3\textwidth}}
\hline
\textbf{Snippet} & \textbf{GPT-3} & \textbf{\gptthreecf} \\
\hline
\begin{tabular}[c]{@{}p{0.3\textwidth}@{}}
\textbf{DR}: Thank you so much for sharing. Are you on any regular medications for that? \\ \textbf{PT}: My only regular medication is birth control -Apri. Low dosage.
\end{tabular} & On birth control. & Only regular medication is birth control -{\bf Apri}. {\bf Low dosage}. \\
\hline
\begin{tabular}[c]{@{}p{0.3\textwidth}@{}}
\textbf{DR}: You had mentioned the headache starting after stopping the medications. Have you had similar headache while stopping medications before? \\ \textbf{PT}: Yes \\ \textbf{PT}: Well that's a complicated question \\ \textbf{PT}: I'm not really sure \\ \textbf{DR}: Okay, no worries. Wanted to know a bit more since you had mentioned about them. That's all.
\end{tabular} & Yes. Stopping medications before. & Yes, has {\bf headache} while stopping medications before. {\bf Is not sure.} \\
\hline
\begin{tabular}[c]{@{}p{0.3\textwidth}@{}}
\textbf{DR}: Do you have pain when you notice penile discharge? \\ \textbf{PT}: no i'm not
\end{tabular} & Did not notice  penile discharge. No pain. & {\bf Doesn't have pain when noticing penile discharge.} \\
\hline
\begin{tabular}[c]{@{}p{0.3\textwidth}@{}}
\textbf{DR}: I have a few questions to ask. How long have you been having this vaginal discomfort? \\ \textbf{PT}: only a few days, since like friday or saturday maybe
\end{tabular} & I have a few questions to ask. How long has she been having vaginal discomfort? & Has been {\bf having vaginal discomfort for only a few days, since friday or saturday.} \\
\hline
\end{tabular}

\caption{Input dialogue snippets along with summaries generated by GPT-3 in column 2 and our approach, \gptthreecf, in column 3.}
\label{tab:examples}
\end{table}


\section{Infusing Medical Knowledge in GPT-3 for use as a Data Generator}
\label{sec:gpt3cf}
We are interested in a model that uses only a small amount of human labeled data to learn an effective medical dialogue summarizer. At the same time, we want such a model to be used in a practical practitioner-in-the-loop setting where medical correctness and patient privacy are of paramount importance.

In order to achieve these goals, we propose a two-pronged approach 
\begin{enumerate}
    \item Introduce \textbf{\gptthreecf} where we infuse medical knowledge into GPT-3 and use it within an inner loop to make it effective at medical summarization.
    \item Leverage \gptthreecf as a data generator to obtain a large training set \footnote{Unlike data at inference time, training data is fixed and can be ensured to be privacy protected} to train an in-house medical dialogue summarization model. Such an in-house model can be used at inference time without the practical constraints related to protecting patient privacy that would require full de-identification to be applied in any conversation, if we were to access the GPT-3 service. It also lends itself well to the practitioner-in-the-loop setting.
\end{enumerate}


\subsection{\gptthreecf: Medically-aware ensemble of GPT-3}
\label{subsec:gpt3ens}
As discussed in \ref{sec:background}, GPT-3 is quite sensitive to the priming context. While one approach may be to provide GPT-3 with the most informative context for a task, this itself is a daunting task and can potentially be tackled if we had a large number of labeled examples (which is the exact problem we want to tackle with GPT-3).

Drawing on the learning from vast literature in ensembling techniques {\it c.f.} \cite{bishop1995neural}, our first key insight is that if we can generate multiple summaries from GPT-3 using a variety of priming contexts, then we should be able to ensemble these outputs to identify the summary that is ideal for the dialogue. This insight leads to a question on how to ensemble multiple text summaries. The answer to this question relies on the core requirement for medical summarization: we care about the coverage of medical concepts mentioned and therefore the best ensembling function is the one that returns the summary with the most medical information in the dialog input. 

In Algorithm~\ref{alg:gpt3-cf} we provide our approach to the medically aware GPT-3 ensemble {\textbf{\gptthreecf}}. We assume access to a small set of labeled examples $\mathcal{L}$. For each input dialog snippet, $T$, we get $K$ summaries, by invoking GPT-3 each time with $N$ examples sampled randomly without replacement from $\mathcal{L}$. We also assume access to a medical entity extractor that can discern the medical concepts from both the dialogue snippet and the summary. The algorithm returns the best summary that has the highest recall in terms of capturing the medical concepts in the dialogue. For this purpose, we use an in-house medical concept extractor \textsc{MedicalEntityRecognizer} that can identify medical concepts from a given piece of text. This extractor has access to the universe of medical concepts based on Unified Medical Knowledge Systems \footnote{\url{https://www.nlm.nih.gov/research/umls/index.html}}, which includes patient symptoms, disorders, laboratory tests and medications. Note that any medical entity recognizer ({\it cf.} \cite{concept19} and references therein) that has coverage for all these types of medical concepts found in medical conversations can be used.

\begin{algorithm}
  \caption{Medically aware GPT-3 ensemble summarizer (\textbf{\gptthreecf}) }
  \label{alg:gpt3-cf}
  \begin{algorithmic}[1]
    \Require{dialogue snippet $T$,  ensembling trials $K$, universe $\mathcal{L}$ of labeled examples, medical entity extractor \texttt{MedicalEntityRecognizer}, \texttt{GPT3}} 
      \State $C^* \gets \texttt{MedicalEntityRecognizer}(T)$
      \For{$i \gets 1, \cdots, K$} 
        \State $S \gets \text{sample $N$ examples from $\mathcal{L}$}$
        \State $\texttt{summary}_i \gets \texttt{GPT3}(S, T)$
        \State $C_i \gets \texttt{MedicalEntityRecognizer}(\texttt{summary}_i)$
      \EndFor
      \State $\texttt{summary}_\texttt{best} \gets \texttt{summary}_{\argmax_{i} \frac{|C_i \cap C^*|}{|C^*|}} $
      \State \Return $\texttt{summary}_{best}$
  \end{algorithmic}
\end{algorithm}



\begin{figure}[ht!]
    \centering
    \includegraphics[scale=0.5]{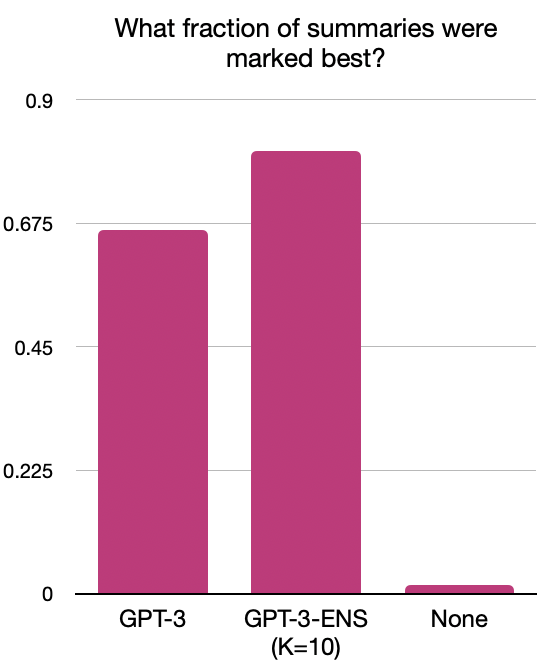}
\caption{Doctor evaluation of which among \gptthree and \gptthreecf summaries they considered ``best'' showing that \gptthreecf is a better approach for labeling}
\label{fig:gpt3_best}
\end{figure}

Reconsider Table~\ref{tab:examples} for qualitative comparison between \gptthree and \gptthreecf.  We can see that summaries obtained using \gptthreecf capture the medical concepts comprehensively (shown in bold) and also have better grammatical structure. We also quantitatively validate the summaries on a small data set distinct from what is used for priming(see \S ~\ref{sec:doceval} for guidelines). In \autoref{fig:gpt3_best}, based on doctor evaluation, we can see that \gptthreecf is significantly better at summarization than \gptthree.

\subsection{\gptthreecf as a data labeler}
\label{subsec:gpt3-data-sim}
We use \gptthreecf described in \ref{subsec:gpt3ens} as our 
labeled data generator. In particular, we use our approach to collect a large amount of labeled examples that serve as inputs to training an off-the-shelf summarization model. This resolves the concern of using GPT-3 in a real world application where the patient's conversation (in its raw form) needs to be exchanged with an external third party such as OpenAI/GPT-3 which may not have design/privacy regulations around HIPAA. In our approach, however, with the help of experts, it is easy to ensure that the dialogues that will used for priming as well as in the training set are chosen following privacy protocols.

\section{Datasets}
\label{sec:datasets}
We collected a random subset of medical conversation dialogues from our chat-based telemedicine platform. Often medical conversation follows a linear ordering of medical history gathering (understanding patient symptoms) that enables creating the summary of the dialog by stitching together summaries of the snippets in chronological order ~\cite{joshi2020dr}. Therefore, we split each dialogue into a series of local dialogue snippets using a simple heuristic: the turns between two subsequent questions by a physician corresponds to a snippet. The length of these snippets ranged anywhere from two turns (a physician question and patient response) to ten turns.

We had medical doctors\footnote{These are the same doctors who practice on the same telemedicine platform.} summarize these snippets. The doctors were asked to summarize the sections as they would for a typical clinical note by including all of the relevant history taking information. If a local snippet did not contain any history taking information it was excluded from annotations. For example in the beginning or end of conversations there may be turns that are purely greetings and not part of the patient history taking process. Further some snippets maybe purely educational in nature and are excluded as well. We eventually obtained a total of 6900 labeled snippet-summary pairs. 
\\

\noindent {\bf Human labeled dataset train/test split:} 
From the 6900 labeled snippet-summary pairs (denoted as $H_{6900}$), we generated a randomly sampled test set $T=500$ that we use in all our evaluations.

The dataset $H_{6900}-T$ is used to generate the priming dataset for GPT-3 related models as well as the datasets we use to train our summarization models.
\\

\noindent {\bf \gptthreecf dataset}: Let \textbf{GCF}$_p^{k}$ be the  dataset of size $p$ generated using \gptthreecf with $k$ ensembling trials. To generate dataset GCF$^{K=k}$, we require $\{\text{H}_n\}_{i=1}^k$ datasets (note the independence on $p$), and thus $n \times k$ labeled examples for priming. These $n \times k$ examples are randomly sampled from the universe of human labeled examples $H_{6900}-T$. In our experiments, we sample without replacement so that no examples are reused across the $k$ tries. To allow comparison between our experiments with different $K$ values, we use the same seed for random sampling.

\section{Evaluation Metrics}
\label{sec:eval}
Multiple studies have shown that automated metrics in NLP do not always correlate well to human judgments as they may not fully capture coherent sentence structure and semantics \cite{fair, salesforcemetrics}. Since medical dialogue summarization would be used to assist health care, it is important for doctors to evaluate the quality of the output.

\subsection{Automated metrics}
While we measure model performance on standard metrics of ROUGE \cite{lin-2004-rouge} \footnote{\raggedright We use the following package with default configuration: \url{https://github.com/google-research/google-research/tree/master/rouge}}, we also measure a model's effectiveness in capturing the  medical concepts that are of importance, and their negations \cite{joshi2020dr}

\noindent {\bf Medical Concept Coverage}: The concept coverage set of metrics captures the coverage of medical terms in the model's output summary with respect to the ground truth. In particular, let $\mathcal{C}$  be the set of medical concepts in the reference summary and $\hat{\mathcal{C}} $ be the set of concepts in the summary output by the model. Then,
$\textrm{Concept recall} = \frac{\sum_{n=1}^{N}  |\hat{\mathcal{C}}^{(n)}  \cap  \mathcal{C}^{(n)}| }  {\sum_{n=1}^{N}|\mathcal{C}^{(n)}|} $ and $\textrm{Concept precision} = \frac{\sum_{n=1}^{N}  |\hat{\mathcal{C}}^{(n)}  \cap  \mathcal{C}^{(n)}| }  {\sum_{n=1}^{N}|\hat{\mathcal{C}}^{(n)}|}$. 

We use these to compute a Concept F1\footnote{Note if there are no concepts detected in the snippet and summary by the entity extractor, then a conservative F1 score of 0 is given for that example.\label{footnote f1}} We use an in-house medical entity extractor to extract medical concepts in the summary. Medical concepts in the decoded summary that weren't present in the original conversation would be false positives and vice versa for false negatives.
 
\noindent {\bf Negation Correctness}:  To measure the effectiveness of the model to identify the negated status of medical concepts,
we use Negex \cite{negex09} to determine negated concepts. Of the concepts present in the decoded summary, we evaluate precision and recall on whether the decoded negations were accurate for the decoded concepts and compute a negation F1$^{\ref{footnote f1}}$.

\subsection{Doctor Evaluation}
\label{sec:doceval}
We also had doctors, who serve patients on our telehealth platform, evaluate the summaries produced by the models. Given the local dialogue snippets and the generated summary, we asked them to evaluate the extent to which the summary captured factually correct and medically relevant information from the snippet. Depending on what percentage of the concepts were correctly mentioned in the decoded summary of the provided snippet, the doctors graded the summaries with \emph{All} (100\%), \emph{Most} (at least 75\%), \emph{Some} (at least 1 fact but less than 75\%), \emph{None} (0\%) labels. 

We also formulated a comparison task where given summaries generated by different models and the associated dialogue, they were asked which summary was the "best" from a usability perspective. Usability was defined as whether the summary could stand in as a replacement for reading the dialogue snippet i.e. whether it captures the correct concepts from the snippet and whether the negations are accurate. The doctors had the ability to use ``all'' and ``none'' in this task depending on if all models being compared captured a good summary or if none of them did. 

To avoid bias, the doctors do not know the model that produced the summary in both the experiments. In the comparison task, the summaries were provided in randomized order so that there is no bias in the order of presentation of the summaries.

\section{Experiments and Results}
\label{sec:expt}
\noindent \textbf{Additional models considered}: To evaluate the efficacy of  \textbf{\gptthreecf} as a source of labeled data generator, we considered models with distinct objective functions for abstractive and hybrid (abstractive/extractive) summarization. We used \textbf{\pegasus} \cite{zhang2019pegasus} for abstractive summarization and Dr. Summarize which we denote as \textbf{\twompgen} \cite{joshi2020dr} for extractive summarization.  For \textbf{\twompgen}, we also use their best performing variant (referred as 2M-PGEN  in \cite{joshi2020dr}) which penalizes generator loss and favors extractive copying. \\

\noindent \textbf{Implementation Details}: We used GPT-3 via the API released by OpenAI\footnote{\url{https://beta.openai.com/}}. Maximum response length was set to 128 tokens, temperature to 0.6 and presence and frequency penalties both set to 0. For \gptthreecf, we use $K=10$ ensembling trials for all our experiments, unless otherwise specified. We observed that $N=21$ was the maximum number of examples we could prime GPT-3 with given the maximum context window length of 2048 tokens for the API. We therefore fix the size of our priming dataset to be 21 in all experiments which invoke GPT-3. Hence we set $L$ to be a random subset of 210 examples from $H_{6900} - T$. 

We followed parameter settings for \twompgen from \cite{joshi2020dr} for pretraining on the CNN-Dailymail dataset. We then fine-tuned on our summarization task dataset with a batch size of 16, source\_max\_tokens = 400, response\_max\_tokens = 200 and max\_grad\_norm clipped at 2.0, for two epochs with a learning rate of 0.15 using Adagrad optimizer.

We used the \pegasus implementation that is pretrained on CNN-Dailymail\footnote{\url{https://huggingface.co/google/pegasus-cnn_dailymail}} provided by \cite{wolf-etal-2020-transformers}. We fine-tuned it on our summarization task dataset with an effective batch size of 256, source\_max\_tokens = 512, response\_max\_tokens = 128 for two epochs using Adafactor\footnote{\url{https://huggingface.co/transformers/main_classes/optimizer_schedules.html\#adafactor-pytorch}} optimizer at the default settings in Hugging Face. For both \pegasus and \twompgen, we used a beam size of four for decoding.

\subsection{Training summarization models using data labeled by \gptthreecf}
\label{subsec:sim-train}
We compare \pegasus and \twompgen trained on human labeled data H$_{6400}$ and \gptthreecf synthesized data GCF$^{K=10}_{6400}$. Note that synthesizing GCF$^{K=10}_{6400}$ needed all of $21 \cdot 10 = 210$ human labeled examples, where $21$, as a reminder, is the maximum number of inputs that can be used for priming. 

Table~\ref{tab:results_models} compares quantitative performance of \pegasus and \twompgen trained on these two datasets. The main observation is that with only 210 human labeled examples, our approach \gptthreecf is able to generate a large amount of training data for both pre-trained summarization models, \pegasus and \twompgen, in such a manner that they yield comparable (or better perfomance) than if they had been trained with only 6400$(\sim$30x) human labeled examples.

For \pegasus, the summarization performance improves drastically compared to model fine-tuned using only the human labeled data. We hypothesize that data generated from \gptthreecf can serve as quality training data for abstractive models such as \pegasus but not so much for hybrid models such as \twompgen due to GPT-3 being a generative language model. The summaries written by our human doctors have writing structure similar to that of a hybrid summarization model such as {\twompgen} that is more extractive in nature. This can explain why \twompgen did not show performance gain when using generated data from \gptthreecf. The key, however, is that it still did perform {\it on par}.


\begin{table}
\renewcommand{\arraystretch}{1.2}
\begin{center}
\begin{adjustbox}{width=0.7\columnwidth}
\begin{tabular}{c|c|c|c|c}
\hline
\multirow{2}{*}{\textbf{Models}} & \multirow{2}{*}{\begin{tabular}[c]{@{}c@{}}\textbf{Train Data}\\\textbf{Source}\end{tabular}} & \multicolumn{3}{c}{\textbf{Metrics}} \\
\cline{3-5}
 & & \begin{tabular}[c]{@{}c@{}}\textbf{Negation}\\\textbf{F1}\end{tabular} & \begin{tabular}[c]{@{}c@{}}\textbf{Concept}\\\textbf{F1}\end{tabular} & \begin{tabular}[c]{@{}c@{}}\textbf{ROUGE-L}\\\textbf{F1}\end{tabular} \\
\hline
\multirow{4}{*}{\pegasus} & $H_{6400}$ & 21.09 & 35.96 & 55.59 \\
& $GCF_{6400}^{k=10}$ & \textbf{28.89} & 40.02 & 53.43 \\
& $GCF_{12800}^{k=10}$ & 26.70 & 40.21 & 56.66 \\
& $GCF_{25600}^{k=10}$ & 28.61 & \textbf{40.58} & \textbf{58.44} \\
\hline
\multirow{4}{*}{\twompgen} &$H_{6400}$ & \textbf{26.75} & 39.95 & \textbf{52.70} \\
& GCF$_{6400}^{k=10}$ & 24.29 & 37.55 & 48.47 \\
& GCF$_{12800}^{k=10}$ & 26.66 & 38.49 & 49.18 \\
& GCF$_{25600}^{k=10}$ & 26.08 & \textbf{39.47} & 50.85 \\
\hline
\end{tabular}
\end{adjustbox}
\end{center}
\caption{Automated evaluation of summarization models trained with different data labeling methodologies.  Note that the amount of human labeled data is still pretty low (210), compared to 6400 when we do not use our approach.}
\label{tab:results_models}
\end{table}



In the same Table~\ref{tab:results_models}, we also present the results with increased amounts of data (12800 and 25600) from \gptthreecf. There is little or no further improvement in the automated metrics of concept and negation F1. However, ROUGE-L F1 improves reflecting the improvements in coherency of the summaries. We leave this area as future work to explore.

\subsection{Effect of combining human labeled data with data labeled by \textbf{\gptthreecf}}
\label{subsec:mix-train}

Since GPT-3 relies on limited local priming context ($N=21$) it may not be agile in providing robust summaries for a multitude of variations in snippets, focusing on the exploitation part of the exploration-exploitation trade-off. We hypothesize that best summaries then will be synthesized by a model trained on a dataset with human and \gptthreecf labeled examples. To evaluate this, we introduced a mixing parameter $\alpha$, the ratio of \gptthreecf labeled examples to human labeled examples. For instance, with 6400 human labeled examples, $\alpha=0.5$ implies the dataset contains 6400 human labeled examples along with $0.5*6400=3200$ \gptthreecf generated examples. We experiment with $\alpha=0.5, 1, 2, 3$. 

 From \autoref{tab:results_mix}, we observe that for both \pegasus and \twompgen, mixture of human labeled and GPT-3-ENS data consistently improves almost all automated metrics for all $\alpha$ values\footnote{Note here that the claim is not that increasing $\alpha$ improves metrics but that mixing GPT-3-ENS and human labeled data improves metrics over models trained only using human data. We leave it as a future work on how to trade-off between human and GPT-3-ENS labeled data.} 
 The lift in metrics is lower for \twompgen, again illustrating the idea we highlighted in \autoref{subsec:sim-train} of GPT-3-ENS data being more amenable to abstractive models such as \pegasus than for hybrid or extractive-biased models such as \twompgen. \autoref{tab:examples_mix} provides qualitative comparison between summaries generated by each of these models. 
 
\begin{figure}
    \centering
    \includegraphics[scale=0.5]{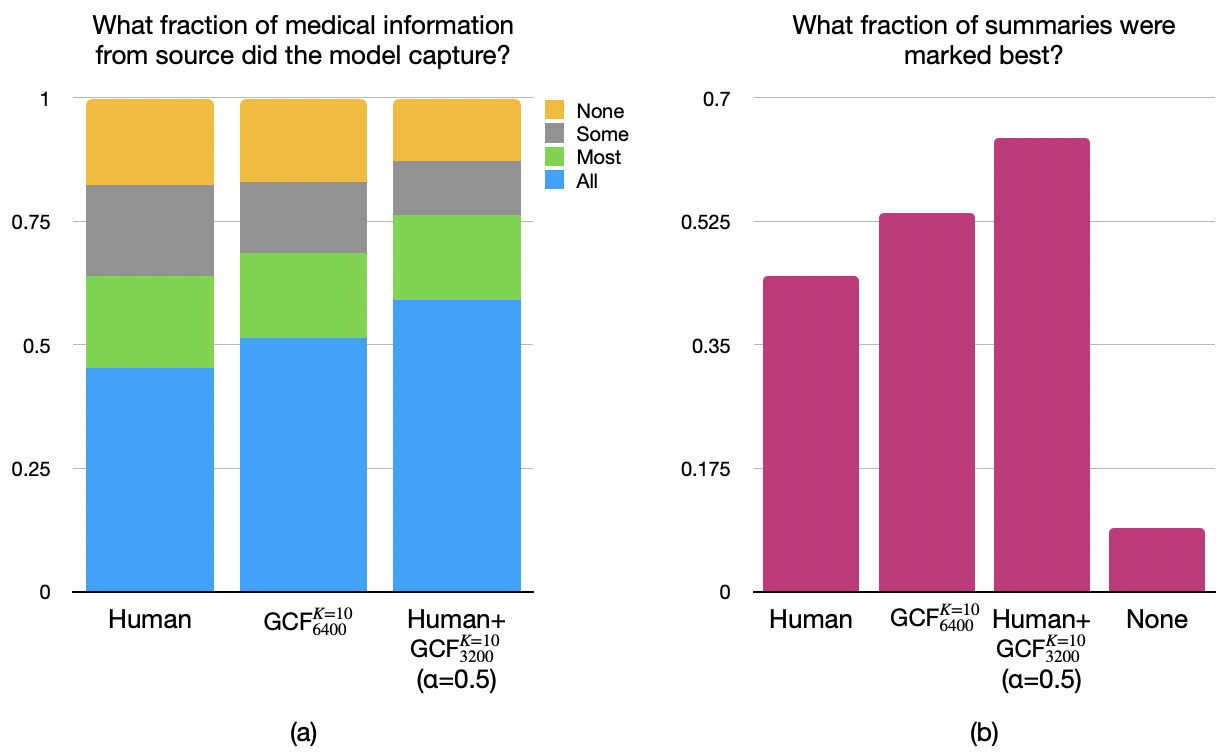}
\caption{Doctor evaluation of amount of medical information covered by summaries provided by \pegasus models and which ones they considered ``best''}
\label{fig:pegasus_human_eval}
\end{figure}

\begin{figure}
    \centering
    \includegraphics[scale=0.5]{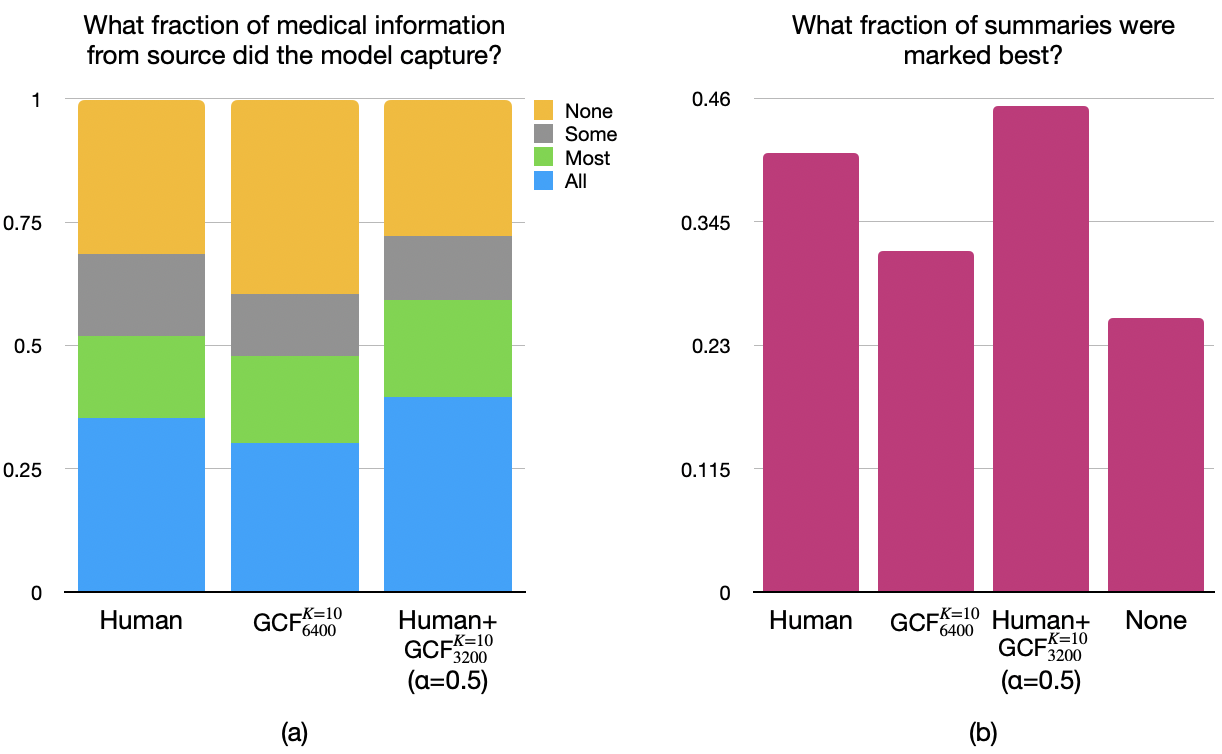}
\caption{Doctor evaluation of amount of medical information covered by summaries provided by \twompgen models and which ones they considered ``best''}
\label{fig:drsum_human_eval}
\end{figure}

 For simplicity, we chose the smallest GPT-3-ENS mix i.e. $\alpha=0.5$ for human evaluation where we ask doctors to evaluate summaries from model trained on human, \gptthreecf and human+\gptthreecf data. \autoref{fig:pegasus_human_eval} and \autoref{fig:drsum_human_eval} show that doctors prefer summaries from the model trained on the mixture data over those produced by models trained on human or GPT-3-ENS data alone, in terms of amount of medical information captured as well as the overall quality of the summary. Furthermore, \autoref{fig:pegasus_human_eval}(b) also shows that for \pegasus, doctors prefer the summaries from a model trained on GCF$^{K=10}_{6400}$ (which needed \textit{only 210 human labeled examples}) over those produced by a model trained on 6400 human labeled examples.

\begin{table*}
\footnotesize
\renewcommand{\arraystretch}{1.3}
\begin{tabular}{p{0.22\textwidth}m{0.22\textwidth}m{0.22\textwidth}p{0.22\textwidth}}
\hline
\textbf{Snippet} & \textbf{Model trained on H$_{6400}$} & \textbf{Model trained on GCF$^{K=10}_{6400}$} & \textbf{Model trained on H$_{6400}$+GCF$^{K=10}_{3200}$} \\
\hline
\begin{tabular}[c]{@{}p{0.22\textwidth}@{}}
\textbf{DR}: Have you ever been tested for any underlying health conditions such as diabetes, hypothyroidism or polycystic ovarian syndrome? \\ \textbf{PT}: No \\ \textbf{PT}: I have been told I have prediabetes
\end{tabular} & Has not been tested for any underlying health conditions. & Hasn't tested for any underlying health conditions such as diabetes, hypothyroidism or polycystic ovarian syndrome & Has not been tested for any underlying health conditions. Has been told has prediabetes.\\
\hline
\begin{tabular}[c]{@{}p{0.22\textwidth}@{}}
\textbf{DR}: Do you have pus appearing discharge from the site? \\ \textbf{PT}: Yes. If the bubbles pop it leaks out a watery substance
\end{tabular} & Has pus appearing from the site. & Pus appearing from the site & Pus discharge from the site. If bubbles pop it leaks out a substance.\\
\hline
\end{tabular}
\caption{Input conversation snippets along with summaries generated by models trained on different data}
\label{tab:examples_mix}
\end{table*}
 






\begin{table}
\renewcommand{\arraystretch}{1.2}
\begin{center}
\begin{adjustbox}{width=0.7\columnwidth}
\begin{tabular}{c|c|c|c|c}
\hline
\multirow{2}{*}{\textbf{Models}} & \multirow{2}{*}{\begin{tabular}[c]{@{}c@{}}\textbf{Train Data}\\\textbf{Source}\end{tabular}} & \multicolumn{3}{c}{\textbf{Metrics}} \\
\cline{3-5}
 & & \begin{tabular}[c]{@{}c@{}}\textbf{Negation}\\\textbf{F1}\end{tabular} & \begin{tabular}[c]{@{}c@{}}\textbf{Concept}\\\textbf{F1}\end{tabular} & \begin{tabular}[c]{@{}c@{}}\textbf{ROUGE-L}\\\textbf{F1}\end{tabular} \\
\hline
\multirow{1}{*}{\pegasus} & $H_{6400}$ & 21.09 & 35.96 & 55.59 \\
$\alpha=0.5$ & \begin{tabular}[c]{@{}c@{}} $H_{6400} + GCF_{3200}^{K=10}$ \end{tabular} & 30.14 & 43.49 & \textbf{62.45} \\
$\alpha=1$ & \begin{tabular}[c]{@{}c@{}} $H_{6400} + GCF_{6400}^{K=10}$\end{tabular} & 30.70 & 43.73 & 60.63 \\
$\alpha=2$ & \begin{tabular}[c]{@{}c@{}} $H_{6400} + GCF_{12800}^{K=10}$\end{tabular} & 29.43 & 41.02 & 59.85 \\
$\alpha=3$ & \begin{tabular}[c]{@{}c@{}} $H_{6400} + GCF_{25600}^{K=10}$\end{tabular} & \textbf{31.93} & \textbf{44.68} & 61.05 \\
\hline
\multirow{1}{*}{\twompgen} &$H_{6400}$ & 26.75 & 39.95 & 52.70 \\
$\alpha=0.5$ & \begin{tabular}[c]{@{}c@{}}$H_{6400} + GCF_{3200}^{K=10}$\end{tabular} & \textbf{27.51} & 40.46 & \textbf{53.39} \\
$\alpha=1$ & \begin{tabular}[c]{@{}c@{}}$H_{6400} + GCF_{6400}^{K=10}$\end{tabular} & 27.18 & 40.36 & 51.00 \\
$\alpha=2$ & \begin{tabular}[c]{@{}c@{}}$H_{6400} + GCF_{12800}^{K=10}$\end{tabular} & 27.19 & \textbf{40.68} & 53.07 \\
$\alpha=3$ & \begin{tabular}[c]{@{}c@{}}$H_{6400} + GCF_{25600}^{K=10}$\end{tabular} & 26.33 & 39.89 & 52.29 \\

\hline
\end{tabular}
\end{adjustbox}
\end{center}
\caption{Combining human labeled datasets with datasets generated using our proposed approach}
\label{tab:results_mix}
\end{table}

\section{Conclusion}
\label{sec:conclusion}
We introduced a medically-aware GPT-3 data labeler, \gptthreecf,  for  the task of medical conversation summarization. At the heart of the approach is a medically aware ensembling criterion that ensembles multiple summaries for an input from a powerful low-shot learner such as GPT-3.  We showed that this approach can generate quality training data for medical dialogue summarization models while ensuring medical correctness. We show that using a very small number of human labeled examples, 210, we are able to produce more medically correct and better quality summaries than using roughly thirty times as many human labeled examples for two different summarization models. In this work we used a simple ensembling technique that dialogue summaries should retain all the medical information discussed in the dialogue. 

\textbf{Limitations and Future Work} In the future, we would like to improve our ensembling function to take into account other medical priors such as affirmations and importance/relevance of the information in the dialog or even come up with a different way of ensembling altogether. Another active area is to understand and model trade-off between the amount of human and GPT-3-ENS labeled data in the training dataset. Thirdly, we can investigate ways to prime GPT-3 in an unsupervised manner that leads to better coverage of latent space of the domain. Lastly, we would also like to investigate other applications in healthcare where the idea of using large pretrained models to generate synthetic labels can be useful.



\bibliography{bibliography}

\clearpage
\appendix
\section{GPT\text{-}3 Prompt}

We utilize a fairly simple prompt to have GPT-3 generate summaries. Each example (snippet\_text, summary\_text) is concatenated to the empty string with the following transformation:"\{snippet\_text\}{[}SUMMARIZED{]}\{summary\_text\}{[}STOP{]}" to form the prompt. We seperate the conversational turns in snippet\_text with the "{[}SEP{]}" token. \autoref{tab:prompt} shows a prompt that would be generated and used to prime GPT-3 given two examples. As mentioned in \autoref{sec:expt} in our experiments we use 21 examples to generate a prompt.

\begin{table}[]
\resizebox{1\textwidth}{!}{%
\begin{tabular}{@{}|l|l|l|@{}}
\toprule
\multicolumn{1}{|c|}{\textbf{Snippet}} &
  \multicolumn{1}{c|}{\textbf{Summary}} &
  \multicolumn{1}{c|}{\textbf{Prompt}} \\ \midrule
\begin{tabular}[c]{@{}l@{}}\textbf{PT:} Today spit out a bit of\\ mucus and noticed a bit of\\ blood.\\ \textbf{DR:} Okay, how long have\\ you been on these medications?\\ \textbf{PT:} About 2 years\end{tabular} &
  \begin{tabular}[c]{@{}l@{}}Has been on these medications\\  for about 2 years.\end{tabular} &
  \begin{tabular}[c]{@{}l@{}}Today spit out a bit of mucus and noticed a\\ bit of blood.{[}STOP{]} Okay, how long have you\\ been on these medications?{[}SEP{]} About 2 years\\ {[}SUMMARIZED{]} Has been on these medications\\  for about 2 years.{[}STOP{]}\end{tabular} \\ \cmidrule(r){1-2}
\begin{tabular}[c]{@{}l@{}}\textbf{DR:} Is the bleeding\\ from the anal opening and\\ not the vagina? Has something\\  similar happened before?\\ \textbf{PT:} yes from the anal opening\end{tabular} &
  \begin{tabular}[c]{@{}l@{}}The bleeding is from the\\ anal opening.\end{tabular} &
  \begin{tabular}[c]{@{}l@{}}Is the bleeding from the anal opening and not the\\ vagina? Has something similar happened before?\\ {[}SEP{]}yes from the anal opening{[}SUMMARIZED{]}\\ The bleeding is from the anal opening.{[}STOP{]}\end{tabular} \\ \bottomrule
\end{tabular}%
}
\caption{Prompt for GPT-3 given two examples}
\label{tab:prompt}
\end{table}

\end{document}